\documentclass[sigconf]{acmart}
\usepackage{todonotes}
\AtBeginDocument{%
  \providecommand\BibTeX{{%
    \normalfont B\kern-0.5em{\scshape i\kern-0.25em b}\kern-0.8em\TeX}}}

\copyrightyear{2020} 
\acmYear{2020} 
\setcopyright{acmlicensed}\acmConference[FDG '20]{International Conference on the Foundations of Digital Games}{September 15--18, 2020}{Bugibba, Malta}
\acmBooktitle{International Conference on the Foundations of Digital Games (FDG '20), September 15--18, 2020, Bugibba, Malta}
\acmPrice{15.00}
\acmDOI{10.1145/3402942.3402951}
\acmISBN{978-1-4503-8807-8/20/09}

\begin{document}

\title{Player-Centered AI for Automatic Game Personalization: Open Problems}

\author{Jichen Zhu}
\email{jichen.zhu@gmail.com}
\affiliation{%
  \institution{Drexel University}
  \city{Philadelphia}
  \state{PA}
  \postcode{19104}}
\author{Santiago Onta\~n\'on}
\authornote{Currently at Google.}
\email{so367@drexel.edu}
\affiliation{%
  \institution{Drexel University}
  \city{Philadelphia}
  \state{PA}
  \postcode{19104}
}

\renewcommand{\shortauthors}{Zhu and Onta\~n\'on}

\begin{abstract}
Computer games represent an ideal research domain for the next generation of personalized digital applications. This paper presents a player-centered framework of AI for game personalization, complementary to the commonly used system-centered approaches. Built on the Structure of Actions theory, the paper maps out the current landscape of game personalization research and identifies eight open problems that need further investigation. These problems require deep collaboration between technological advancement and player experience design.
\end{abstract}



\maketitle

\section{Introduction}

Automated personalization is becoming an integral part of everyday life. We consume products suggested to us by recommendation systems. We find out what is going on with friends and in the world through content feeds curated for us individually. More and more, our games and digital apps can figure out our needs and preferences and adapt accordingly. In education, adaptive technology allows students to learn at their own speed~\cite{wang2017interactive}. 
Personalization is not limited to the digital world; it is also transforming the manufacturing industry~\cite{tseng2010design,kumar2007mass}. These advances propel the business world to anticipate the next evolution: the Internet of Me, where mass personalization is driven by users' individual characteristics such as biology~\cite{Anderson2018,Lardinois2013,kumar2007mass}. 



A significant amount of research has been devoted to automatic personalization in digital applications, especially in Internet applications \cite{Churchill2013}. As the content of the Internet services grows, personalized applications such as recommendation systems help to mitigate information overload and decision fatigue \cite{Churchill2013}.  This body of work ranges from relatively simple changes on a web page (e.g., using the name of each user) to complex customization using deeper models of user needs and behavior \cite{Kramer2000}. 

Computer games are a relatively new domain for personalization. Compared to the classic personalization domains of information seeking and e-commerce, people play games for a broader range of reasons (e.g., challenge, exploration, aesthetic experience, and social activity). It is hence more difficult to identify individual players' needs and preferences that the games should adapt to. Besides, computer games generally involve more complex content and user interaction than other digital applications such as websites. Typical gameplay is multi-sensory (e.g., visual, auditory, and tactile) and contains multiple layers of meanings (e.g., formal rules and stories). To personalize games thus requires further technological advancement (in how to procedurally adapt more complex game content) and new design principles (of how to personalize for various player needs) than what we have learned from the classic domains of personalization. Computer games, therefore, are an excellent domain for researching the next stage of personalization technology. In this paper, we adopt Bakkes, Tan, and Pisan's definition \cite{bakkes2012personalised} that personalized games are those games that adapt themselves based on information about the current player, e.g. by determining the difficulty level appropriate to the current player automatically.

This paper's key argument is that existing AI research on personalized games can benefit from more player-centered perspectives. Despite their technical contributions, most existing work in this area is primarily geared towards more sophisticated algorithms and system capabilities. This system-centered approach to personalization has been attempted in early research of personalized web applications. It led to practices that tried ``to find uses for the tools, and deploying the coolest new features'' and made these applications less useful to the population they were supposed to serve~\cite{Kramer2000}. To avoid similar drawbacks, the game AI research community can benefit from further aligning the technical research with deep models of player needs and behavior. 

In this paper, we present our initial work to strengthen the above alignment by mapping the current landscape of game AI research in personalization through a cognitive science theory. In particular, we use Norman's cognitive theory on Stages of Actions \cite{norman2013design} to examine the state of the art research along each stage of what a player goes through and identify open problems for further investigation. Our goal is to present the current state of game AI research on adaptive personalized games that complement technology-centered overviews. The paper's central position is that to advance player-centered AI for game personalization, we need to combine technological advancement and player experience/game design. 

\begin{figure}[t!]
\includegraphics[width=0.9\columnwidth]{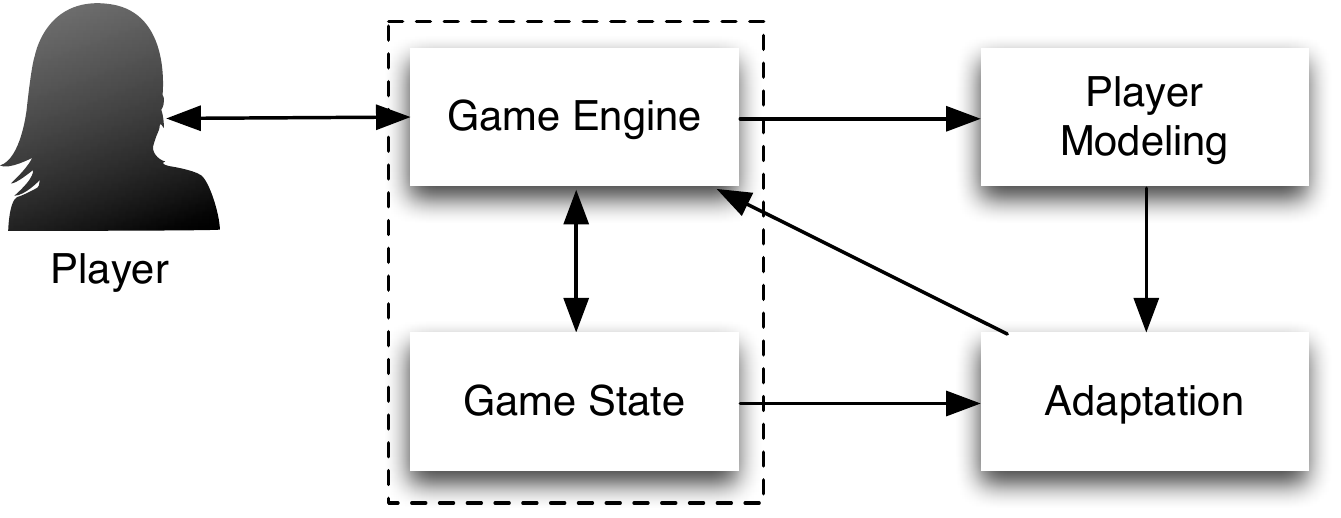}
\centering
\caption{A typical system-centered view to an adaptive personalized game system.}
\label{fig:adaptive-system}
\end{figure}

There have been several overviews on the subject of game personalization. For example, Bakkes, Tan, and Pisan \cite{bakkes2012personalised} offered a taxonomy of personalization in games research. More recently, Snodgrass et al.~\cite{snodgrass2019like}, combining a systematic survey on game design and AI algorithms on personalization, proposed a framework for personalization through the adaptation of the Player, Environment, Agents, and System. Existing related work analyzes personalization from the game developers' point of view (e.g., AI researchers, game designers). By contrast, this paper is among the first attempts to understand game personalization from the vantage view of players' cognitive processes.

\begin{figure}[t!]
\includegraphics[width=0.9\columnwidth]{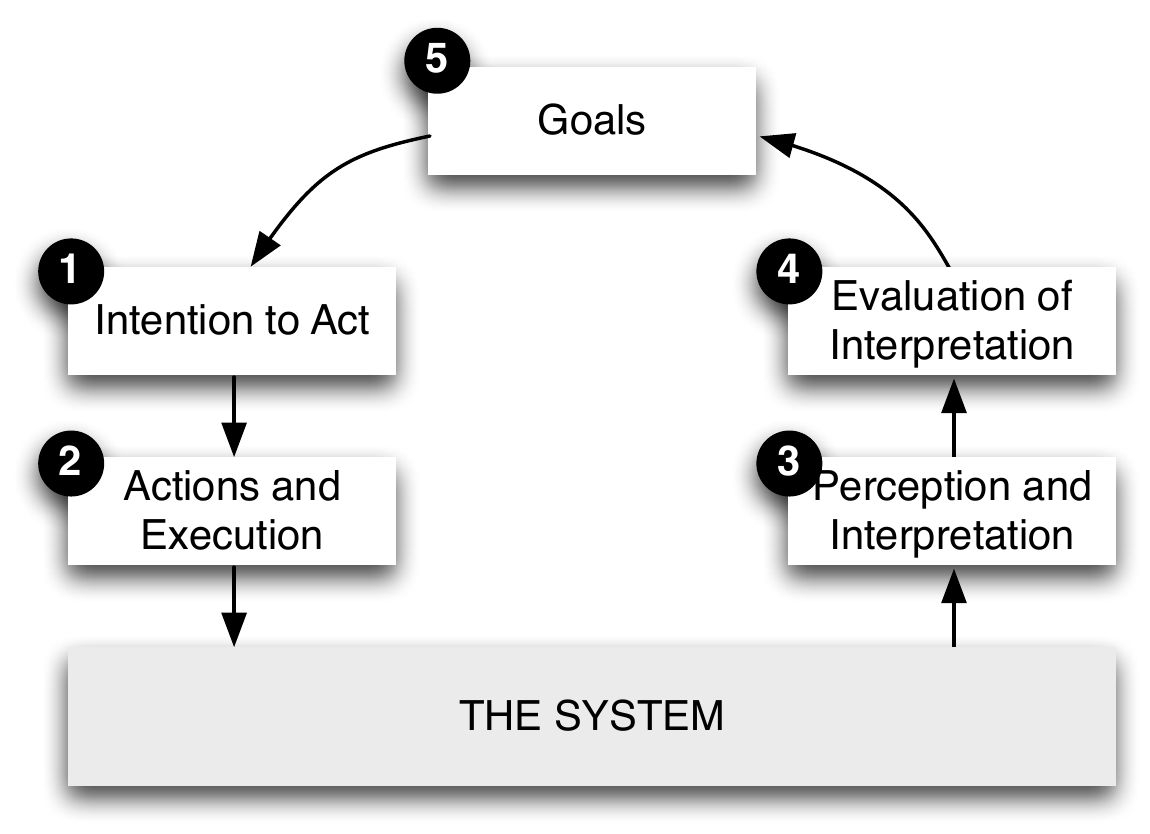}
\centering
\caption{A condensed framework based on Norman's Structure of the Action cycle.~\cite{norman2013design}}
\label{fig:action-cycle}
\end{figure}

\section{Theoretical Background}

The idea of personalization originated as a reaction to mass production. In 1956, the same year as the Dartmouth summer school of AI, Wendell R. Smith coined the term ``market segmentation'' to capture trends in marketing \cite{smith1956product}. He challenged the long-held assumption of heterogeneity in both supply and demand sides of the market. Instead, Smith proposed to segment the market into smaller homogeneous ones to better capture the ``diversity and variations in consumer demand.'' If we take Smith's idea to the extreme, each person becomes a complete segmentation --- personalization.

With few exceptions, game AI publications on adaptive games typically represent adaptive games through system architecture diagrams similar to Figure \ref{fig:adaptive-system}. While this approach to thinking of adaptive games shed light on the connections between different AI components to achieve game personalization, it black-boxes players' needs and requirements. While it is necessary for technological advancements, it risks developing technology for its own sake. 

This paper attempts to examine the current state of game personalization AI techniques through a novel player-centered lens. We use Don Norman's cognitive theory on the structure of actions \cite[p.47]{norman2013design}, which is foundational work in the field of human-computer interaction. It describes the underlying cognitive processes a person goes through when taking action in the world into seven stages: goals, intention to act, actions, executions, perception, interpretation, and evaluation of interpretation. For this paper, we condense the process into five stages, combining actions with execution, and perception and interpretation. We also replace ``the world'' in Norman's original diagram with the focus of our analysis: Adaptive game systems (Figure \ref{fig:action-cycle}. Notice Figure \ref{fig:action-cycle} is the complementary view of Figure \ref{fig:adaptive-system}. 

\begin{figure*}[t!]
\includegraphics[width=\textwidth]{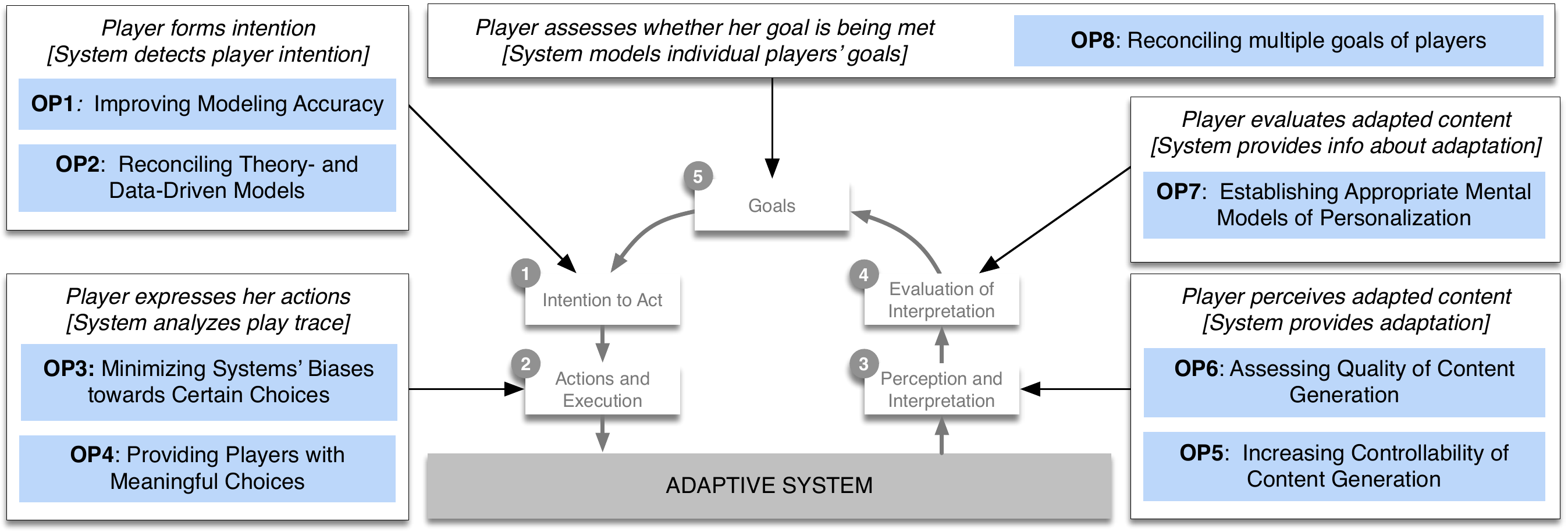}
\centering
\caption{Eight open problems of current personalized game research in relation to the different stages of players' action cycle.}
\label{fig:action-cycle-with-ops}
\end{figure*}

\textbf{Goals} refer to something to be achieved, often vaguely stated, such as ``find a movie to watch.''  A goal is then translated to an \textbf{intention to act} (marked 1 in Fig \ref{fig:action-cycle}). The intention is more specific, such as ``find a good movie on {\em Netflix}''. A goal can be satisfied with different intentions and sequences of actions. In our example, the same goal could also lead to the intention of calling a friend for recommendations. The next stage is \textbf{actions and execution}  (2) where the person specifies an action sequence and executes it --- ``log into {\em Netflix} and browse the {\em Top Picks} for me.'' 

Once actions are taken in the system, the stage of \textbf{perception and interpretation} (3) allows the user to perceive the state of the system and interpret that state (e.g., the list of movies that {\em Netflix} recommends to me). Then the user can form her \textbf{evaluation of interpretation} (4) by asking the question, in our example, ``is there something I like in this list?'' If she determines the outcome of her actions has not satisfied her goal, the user may choose a different intention or modify her goal and go through the cycle again.  

Stages (1) and (2) are about execution: what do we do to the world/system. If the user encounters difficulties in either stage, it creates {\em the gulf of execution} where the user does not know what to do (e.g., when the user cannot find {\em Top Picks} list). Stages (3) and (4) are about evaluation: comparing what happened with what we wanted to happen. Issues in either stage lead to {\em the gulf of evaluation} where the user cannot tell what happened or whether their goal was met (e.g., the system did not respond after pressing the scroll button in {\em Top Picks}).


\section{State of the Art and Open Problems}

According to Norman's theory, the stages of actions are the fundamental cognitive process each person goes through many times when interacting with any system, including adaptive personalized games. This framework sheds light on the gulf of execution and the gulf of evaluation, the underlying causes for the most common usability issues. It also differentiates the stages that are observable by an external spectator (Stages 2 \& 3) from those that are latent and not directly observable (Stages 1, 4 \& 5), highlighting critical challenges for adaptive game AI to model player intent, goals, and appraisal. Therefore, it provides a useful framework for understanding the current state of game personalization research and identifying critical open problems for developing player-centered approaches. 
Figure \ref{fig:action-cycle-with-ops} shows how a player goes through the stages in the action cycle in an adaptive game and how the system functions accordingly. It also highlights the eight open problems discussed in this section.

\subsection{Intention to Act}

Currently, the gap between actual player intent and what an adaptive game ``thinks'' as player intent can be vast. Using algorithms to identify user intent is particularly challenging because the player's intention is not directly observable and therefore has to be inferred based on observable features. Compared to Web applications with comparatively simple user actions, games often feature a more complex set of player choices and player experience. To demonstrate the difficulty of the problem, a player describes his experience of getting lost in one area of {\em Silent Hill: Shattered Memories}, an adaptive horror survival game. The game misunderstood it as an indication of his intent to choose to stay there and hence adapt the game to give the player more related content\footnote{Retrieved from \url{https://www.reddit.com/r/silenthill/comments/3btew0/how_accurate_were_your_psych_profiles_for/} on January 31st, 2020.}. 





A direct approach to predict player intent is through player goal recognition. Researchers have used different machine techniques such as Bayesian models~\cite{Synnaeve2011,albrecht1998bayesian,baikadi2014generalizability}, Markov logic networks~\cite{ha2011goal}, and long short-term memory networks (LSTMs)~\cite{Min2016} for plan, activity, and intent recognition in computer games. These approaches have yielded reasonable results, given the complexity of the problem. 

An indirect approach to approximate intent is player modeling. The underlying assumption is that if we can model certain relevant aspects of the player (e.g., gameplay preference), we can infer their individual intent and thus provide better personalization. Player modeling has been extensively studied. For example, work exists on modeling player types~\cite{bartle1996hearts,heeter2011beyond}, preferences~\cite{thue2008passage,sharma2010drama},  player experience~\cite{pedersen2009modeling}, skill level~\cite{missura2009player,jennings2010polymorph,zook2012temporal} or player behavior~\cite{Drachen2009,valls2015exploring,holmgaard2014evolving,harrison2011using}, among other aspects.
Despite the vast body of literature on this topic, two main open challenges still remain in player modeling.

\begin{figure*}[t!]
\includegraphics[width=0.8\textwidth]{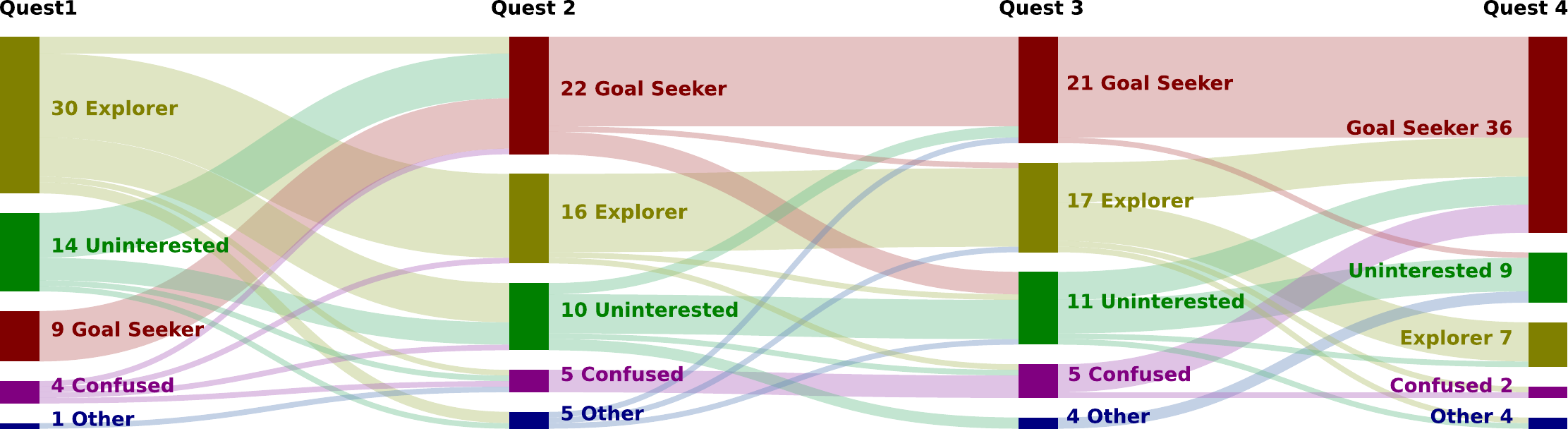}
\centering
\caption{Alluvial diagram of how participants (n=58) shift their observed play styles between the four quests in the {\em Solving the Incognitum} game (reproduced with permission from~\cite{valls2015exploring}).}
\label{fig:taemile}
\end{figure*}

\subsubsection{Open Problem (OP) 1: Improving Modeling Accuracy.} While some of the work cited above shows it is possible to model general aspects of player groups with significant accuracy, modeling aspects of particular players is still a big challenge. One of the main problems is that to model an individual player, a considerable amount of data about that player would be needed. However, even if such data was available, labeling such data is a challenging problem. Even if the researchers could directly observe player behavior for an extended period of time, it is not always clear what the player's intention is at any given time. 

Some approaches, like collaborative filtering~\cite{resnick1994grouplens}, commonly used to model user preferences for product recommendation, avoid the need for labeling data. Collaborative filtering makes predictions about the preferences of an individual by comparing her known preferences to those of other individuals. The assumption is that individuals with similar preferences about a product will likely have the same choices about others.
However, the price they pay is not having models tuned to a specific individual, as preferences of an individual for a given product are assumed to be similar to other individuals who had similar preferences for other products.

\subsubsection{OP 2: Reconciling Theory-Driven and Data-Driven Models.} As reported several times~\cite{yannakakis2013player}, there is a tension between top-down (theory-driven) and bottom-up (data-driven) approaches. Some work on player modeling is based on existing theories, for example, those from psychology (e.g., Self Determination Theory~\cite{bouvier2014trace,sawyer2018modeling}). Others are directly based on the data itself (e.g., ~\cite{valls2015exploring}). For a recent systematic review of data-driven player modeling, readers can refer to \cite{hooshyar2018data}. However, the research community has no consensus on what to do when those approaches disagree. For example, the learning science theory upon which the work of the TAEMILE project was based on assumed that the property being modeled was a trait, but the data showed that it shifted regularly (Figure \ref{fig:taemile}). When this situation arises, it is unclear what is the right path forward: was the theory incorrect, or was the modeling approach used to interpret the data not accurate?



\subsection{Actions and Executions}
Since the AI infers a player's intention based on her observable actions, player modeling accuracy is thus directly tied to the range and quality of the choices available to the player. The quality of choices in this paper refers to the distance between the ``true actions'' a player would like to take and the set of actions at her disposal in the game. In most games, there is a gap between the ``true actions'' and what is offered by the game. This is partly because choices in games are expensive, especially when they involve art assets, animations, and voice acting. Game designers and researchers used creative ways such as false choices~\cite{day2017agency} and fold-back branching structure~\cite{Crawford2004} to increase players' perceived agency without significantly growing game development costs. Another reason that shapes the options available to players is game designers' design intent for player experience. 



Existing research on the quality of player choices mainly focuses on player agency and narrative immersion~\cite{Murray1997,Harrell2009,fendt2012achieving}. However, as a research field, we do not fully understand how to design game choices that can help the adaptive system more accurately predict player goals and other player characteristics. 

\subsubsection{OP 3: Minimizing Systems' Biases towards Certain Choices.}
\begin{figure}[t!]
\includegraphics[width=\columnwidth]{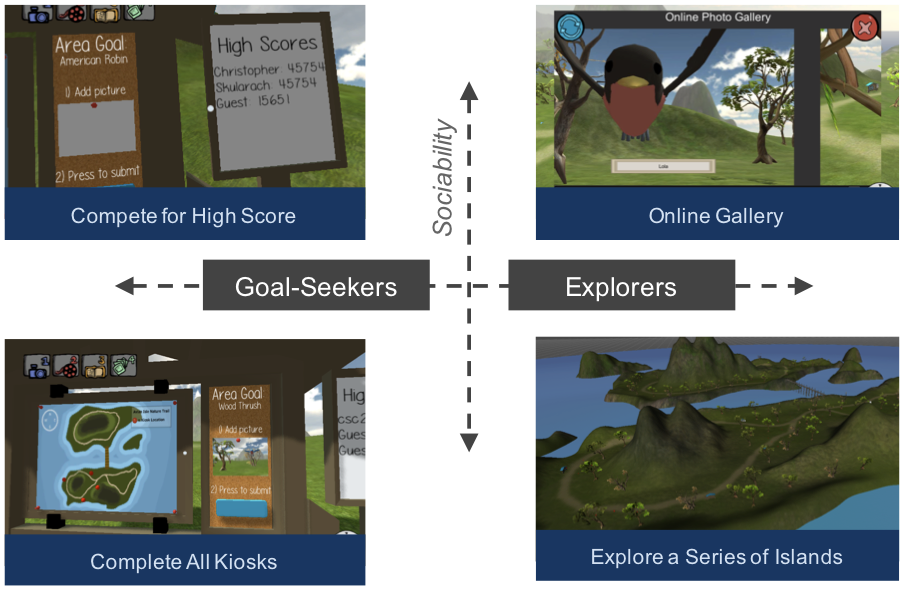}
\centering
\caption{Reward mechanisms for each play style in {\em Avian} (reproduced with permission from~\cite{patterson2014avian}).}
\label{fig:avian}
\end{figure}

The prerequisite of using player choices to infer their intention and preferences is that players can choose, relatively freely, options truly represent them. In other words, they need to be able to express their intent/preferences/other traits with little external influence. Compared to other interactive systems, designing player choices in games represents additional difficulties because of their built-in reward structure. As most players are motivated to win, they may choose options that help them win instead of options that better represent them. For instance, in our {\em Avian} project~\cite{patterson2014avian}, an educational game on citizen science and bird watching, we intended to predict players' play style based on how they play the game. In the first version of the game, we found that most players chose to take actions associated with one playstyle. After looking into it further, we realized players selected these actions because that is how to win the game and where most gameplay content exists. With that revelation, we expanded the game design so that different types of players can find meaningful ways to satisfy their needs while progressing in the game (Figure~\ref{fig:avian}). For explorers, for example, we added more in-games space to discover and an online gallery for those who are social. As a result, we observed a more balanced distribution of different player types. 

This example further illustrates how the success of game AI is heavily tied to game design. When using theory-driven player modeling approaches, it is important to ensure that the different player attributes that the game AI is modeling are sufficiently supported in the player experience. However, when using data-driven approaches, it is not clear how this problem can be addressed, as there is no pre-existing set of categories to inspect during game design before data collection. Further research is needed on the design of choices for effective player modeling. 



\subsubsection{OP 4: Providing Players with Meaningful Choices.}
For a choice to be truly meaningful, what the player chooses should lead to consequential changes. To do so requires a significant amount of game content and risks what is known as the {\em authorial bottleneck problem}~\cite{gervas2009computational} in the context of interactive stories.

Several recent research efforts aim to alleviate this authorial bottleneck. For example, the work of Valls et al. explored how to automatically extract domain knowledge from existing stories written in natural language and then use such domain knowledge to generate new stories~\cite{valls2017towards}, or even game worlds~\cite{valls2013towards}. Another example is by Li et al.~\cite{li2013story} to use crowd-sourcing to obtain plot graphs that can then create interactive experiences. A more recent example is the experimental game {\em AI Dungeon 2}\footnote{\url{https://aidungeon.io}}, where a state of the art language model (GPT-2~\cite{radford2019language}) is trained to generate text based on arbitrary user input in a MUD-style text-based game. Finally, in general, the field of procedural content generation (PCG)~\cite{shaker2016procedural} can also have a significant impact on alleviating this authorial bottleneck.

On a related issue, game designers have long known the dilemma that people who lack choices seem to want them and often will fight for them. Yet at the same time, people with too many options intensely dislike it \cite{schell2008art}. For adaptive games to effectively capture high-quality player choices that can be used for player modeling, we need more work to understand how to increase the efficiency of player choices and avoid decision fatigue. 



\subsection{Perception and Interpretation}
Assume we have a perfect model of the player and the balanced options from which they can select. Now, the player selects her ``true action'', how can now the game adapt in a way that the player can perceive?

Several areas of research have proposed different ideas that contribute to answering this question. For example, the narrative planning community has developed several planning-based approaches to adapt a story based on a player's actions to maximum flexibility, while still ensuring the story advances in the direction desired by the designers. Riedl and Stern~\cite{riedl2006believable} presented the {\em Automated Story Director}, which can detect when the target story goal cannot be reached given the player actions, and re-plan accordingly to achieve the desired narrative effect. This is an instance of what is more generally called {\em drama management} or {\em experience management}~\cite{weyhrauch1997guiding,mateas2002behavior,heeter2011beyond,thue2018toward,zhu2019experience}. Experience Management studies AI systems that automatically adapt interactive experiences such as computer games to serve specific users better and to fulfill specific design goals. For example, experience management techniques have been designed to adapt interactive game experiences to follow the desired story arc~\cite{weyhrauch1997guiding}, or to adjust the difficulty of a game dynamically~\cite{hunicke2005case}. Experience management techniques have been explored for several decades. Although its effectiveness has been demonstrated in commercial deployment in games such as {\em Left 4 Dead 2}, current approaches still have limitations. For example, detecting if the current user is being served, or the goals of the experience manager have been achieved is very challenging, as this might involve automatically detecting, for instance, if the user is engaged, which is an open problem.

All these approaches need to balance {\em player autonomy} (the freedom of the player to behave as they prefer in the game), with {\em authorial intent} (the goal that the game designer had in mind, and that is what the AI adapting the game is aiming for). Example authorial intents include trying to adapt a story to make it conform to a proper Aristotelian arc, or trying to make sure the player is exposed to a pre-specified set of concepts in educational games.

Another promising direction for algorithmic adaptation in games is procedural content generation (PCG). PCG work aimed at adaptive games includes automatic difficulty adjustment via PCG~\cite{jennings2010polymorph}, generating levels specific to player style~\cite{shaker2010towards,togelius2006making,snodgrass2016controllable,valls2017graph}, or the general idea of experience-driven PCG~\cite{yannakakis2011experience}. Currently, existing approaches have their limitations. For instance, in PCG, controllability and quality assessment are two key open challenges.

\subsubsection{OP 5: Increasing Controllability of Content Generation.}
Although there has been some work in this direction (such as some reported above), the general problems of how to algorithmically specify the desired properties of the content we want a PCG algorithm to generate, and how to guide the algorithms to generate it is still unsolved. Additionally, how to exploit player models in general to create personalized content is still an open problem, even though some work exists~\cite{shaker2010towards,togelius2006making,snodgrass2016controllable,valls2017graph,togelius2013controllable}. More generally, how can we generate new adaptations of a game to achieve the desired effect without having to pre-author all of them.  
For example, in the work of Valls et al.~\cite{valls2017graph}, a graph-grammar approach is proposed to generate new puzzles for an educational game about learning parallel and concurrent programming called {\em Parallel}. In this approach, the set of desired concepts the player is supposed to practice (e.g., {\em mutual exclusion}) and the set of concepts the player is not supposed to have mastered yet is given to the PCG component, which can generate a level with those specifications by activating and deactivating rules in the grammar that create problem instances with specific concepts. Although reported results show promise, unanticipated rule interactions can still generate a level that contains an undesired concept.

\subsubsection{OP 6: Assessing Quality of Content Generation.}
A second major problem in this area is how to assess the quality of the generated adaptations/content. Which are the metrics to evaluate this content? How do we avoid ``catastrophic failures''? Some work exists in this direction~\cite{summerville2018expanding,shaker2016evaluating}, trying to propose objective metrics to assess the quality of content, but to date, no automated metric has shown good enough correlation with human reported evaluations~\cite{summerville2017understanding}. 

Using the same example of the {\em Parallel} game above, even if it is possible to automatically identify if a certain concept (e.g., {\em deadlock}) appears in a given puzzle, doing so is NP-complete, as it often requires exploring the set of all possible execution orders of the threads in the level. Additionally, even if we could detect the presence or not of a concept, it is unclear how to assess if the puzzle is a ``good example'' of the concept for pedagogical purposes.

\subsection{Evaluation of Interpretation}

For players to fully evaluate their personalized experience, they need to be able to form the appropriate mental models of whether and how the system adapts to them. In traditional user experience design literature, the doctrine of seamless design \cite{norman2013design,bolter2003windows,swearingen2002interaction,dix2003human} would advocate that players should not be aware that content has been personalized for them. After all, ``a good tool is an invisible tool'' \cite{weiser1994world}. However, this design guideline has been increasingly questioned in the context of adaptive personalized systems. Evidence starts to show that when users are not aware of the adaptive nature of the system, they often cannot utilize it fully. For instance, recent studies show that more than half of {\em Facebook} users are not aware that their news feed is personalized by algorithms~\cite{rader2015understanding}. Even for those who are aware, they can't make sense of how exactly it works. As a consequence, these users build incorrect folk theories that are very different from how the algorithm actually works. This misconception can sometimes negatively impact real-world relationships~\cite{eslami2015always}. 
Furthermore, when users are completely shielded from knowing how the personalization mechanism works, they lose the ability to inspect it. The phenomenon of filter bubbles~\cite{pariser2011filter} is an example of what could go wrong.  



The issue of how and to what extent players should be able to evaluate their personalized gameplay experience has not received sufficient attention. Similarly, in an adaptive user interface, researchers found that when the users are were aware that the personalized interface changed or why it changed, they simply ignored it \cite{Furqan2017}. Another example is the work of Lau~\cite{lau2009programming}, who analyzed a collection of AI-powered adaptive text editors (SMARTedit, Sheepdog, and CoScripter) and why they failed. Out of the five reasons uncovered, one of them was not having ``a model users can understand'', highlighting one more time that helping the user create appropriate mental models of what an adaptive system is doing is vital for its success. Finally, another interesting work in this direction in the context of recommender systems is that of Ekstrand et al. \cite{ekstrand2015letting}, who allowed users to choose which recommendation algorithm to use, thus making them aware of the algorithms. This resulted in 25\% of users using the feature.


%

\subsubsection{OP 7: Establishing Appropriate Mental Models of Personalization.} 
Despite the growing evidence that users need to develop the proper mental model to take full advantage of personalization, this topic has not received enough attention from the game research community. Most overview papers of player modeling~\cite{yannakakis2018artificial,yannakakis2013player,smith2011inclusive} are written from an algorithmic perspective and do not include these player-centered aspects. 
Along with the rising interest in trust, transparency, and ethics of AI \cite{goodman2017european,ribeiro2016should}, the topics such as how to establish the right mental model of personalization and how much information about the personalization mechanism should be revealed to players should be further studied. 


\subsection{Goals}
Personalization aims to model the player in order to changes the player experience accordingly. Simultaneously, personalization is used to influence players and guide their choices and actions. This is also true outside of games, in most apps and websites we use every day. Consider, for example, Amazon and the products it shows you upon entering a search query. Amazon uses user modeling to identify products that might suit your profile, but at the same time, your choices are influenced by the personalization algorithms. Therefore, personalization also works in the other direction: it also changes the player to match a specific digital experience. 

This feedback loop has been referred to as the {\em paradox of personalization}~\cite{koponen2015future}. As a consequence of all previous gaps, and this paradox, and as pointed out by Koponen~\cite{koponen2015future}, ``personalization remains unfulfilling and incomplete. It leaves us with a feeling that it serves someone else's interests better than our own.''
 
\subsubsection{OP 8: Reconciling multiple goals of players.}
Humans are complex and have different co-existing goals and preferences. Most existing work simplifies the problem by focusing on modeling one aspect of player characteristics or player experience. As personalized adaptive games become more sophisticated, another open problem is to algorithmically reconcile multiple goals and preferences of an individual, some of which may be contradictory to one another. 

%
An example of work towards addressing this open problem can be found on the TAEMILE~\cite{valls2015exploring} project mentioned above. TAEMILE is an interactive learning environment for earth science. The game attempts to model goal seeker vs. explorer learning behavior. However, in the context of learning, the player's short-term learning preferences of goal-seeking or exploring often do not match her long-term goal of mastering the subject. This is because an effective pedagogical experience requires both. To cater to the player's long-term needs, TAEMILE did something different from many similar projects: once a player model is built, instead of adapting the game to the player's short-term learning behavior, it is used to nudge the player to try the opposite behavior. In a way, this project uses player modeling to identify the play style of a player, and then use the information to expand how she learns.

As multi-player games become popular among players, further research is also needed to personalize a shared gameplay experience between multiple people. One approach is to expand the framework of experience manager to incorporate models of multiple players~\cite{zhu2019experience}.

\section{Conclusions}
In conclusion, computer games represent an ideal research domain for the next generation of personalized digital applications. To reach the full potential of personalized games, we argue that a player-centered approach for personalization AI is necessary. We hence presented a novel player-centered framework, complementary to the commonly used system-centered approaches, to synthesize existing AI research on personalized games. Built on the cognitive science theory on the structure of actions, we mapped current game personalization research based on the cognitive process a player goes through. As a result, we identified eight open problems that need further investigation. These problems require deep collaboration between AI researchers and game designers.

\begin{acks}
This paper is partially based on Jichen Zhu's invited talk at the 13th AAAI Conference on Artificial Intelligence and Interactive Digital Entertainment (AIIDE'17). This work is partially supported by the National Science Foundation under Grant Number IIS-1816470. 
\end{acks}

\bibliographystyle{ACM-Reference-Format}

\end{document}